\documentclass[11pt]{article}

\usepackage[final]{acl}

\usepackage{times}
\usepackage{latexsym}

\usepackage[T1]{fontenc}

\usepackage[utf8]{inputenc}

\usepackage{times}
\usepackage{latexsym}
\usepackage{amsmath}
\usepackage[table]{xcolor}

\definecolor{myblue}{RGB}{160,235,255}
\newcommand{\graycell}[1]{\textcolor{gray}{#1}}

\usepackage{multirow}
\usepackage[table]{xcolor} 

\usepackage{booktabs}
\usepackage{amssymb}
\usepackage{graphicx}
\title{A Better Spur Should Start From Each Objective}

\author{%
  Shanwen Mao$^{1}$ \And
  Hao Zhang$^{1}$ \And
  Guangtao Nie$^{2}$\thanks{Corresponding author.} \And
  Zhiheng Li$^{3}$ \AND
  Huimu Wang$^{3}$\thanks{Corresponding author.} \And
  Sulong Xu$^{2}$ \And
  Simiu Gu$^{2}$ \AND
  $^{1}$Harbin Institute of Technology, Harbin, China \\
  $^{2}$JD.com Inc., Beijing, China \\
  $^{3}$Institute of Automation, Chinese Academy of Sciences, Beijing, China \\
  \texttt{24s103313@stu.hit.edu.cn, zhh1000@hit.edu.cn} \\
  \texttt{\{nieguangtao1, xusulong, nick.gu\}@jd.com} \\
  \texttt{\{lizhiheng2025, huimu.wang\}@ia.ac.cn}
}

\begin{document}
\maketitle
\begin{abstract}
Real-world Multi-Objective Reinforcement Learning (MORL) often suffers from sparse rewards, reward conflicts, and late-stage reward tug-of-war, causing traditional linear scalarization to experience severe metric oscillations. To address optimization conflicts among multiple objectives in real-world deployment scenarios, we propose Multi-Marginal Preference Optimization (MMPO), a fine-grained framework that intervenes at the data, gradient, and constraint levels rather than relying on coarse-grained global scalarization. Specifically, MMPO performs reward smoothing and exposure debiasing to mitigate sparse and biased rewards, applies priority-aware orthogonal projection to decouple conflicting gradients, and introduces self-prompted gradient constraints to prevent dominant objectives from overwhelming weaker ones. Experiments on real-world e-commerce datasets show that MMPO improves training stability and consistently achieves better performance across conflicting metrics. Moreover, it generalizes robustly to broader tasks such as ToolRL and code generation, demonstrating its effectiveness as a practical paradigm for multi-objective alignment. It is especially applicable to scenarios with multi-objective and sparse rewards.
\end{abstract}

\begin{figure}[t] 
\centering
\includegraphics[width=1.1\columnwidth]{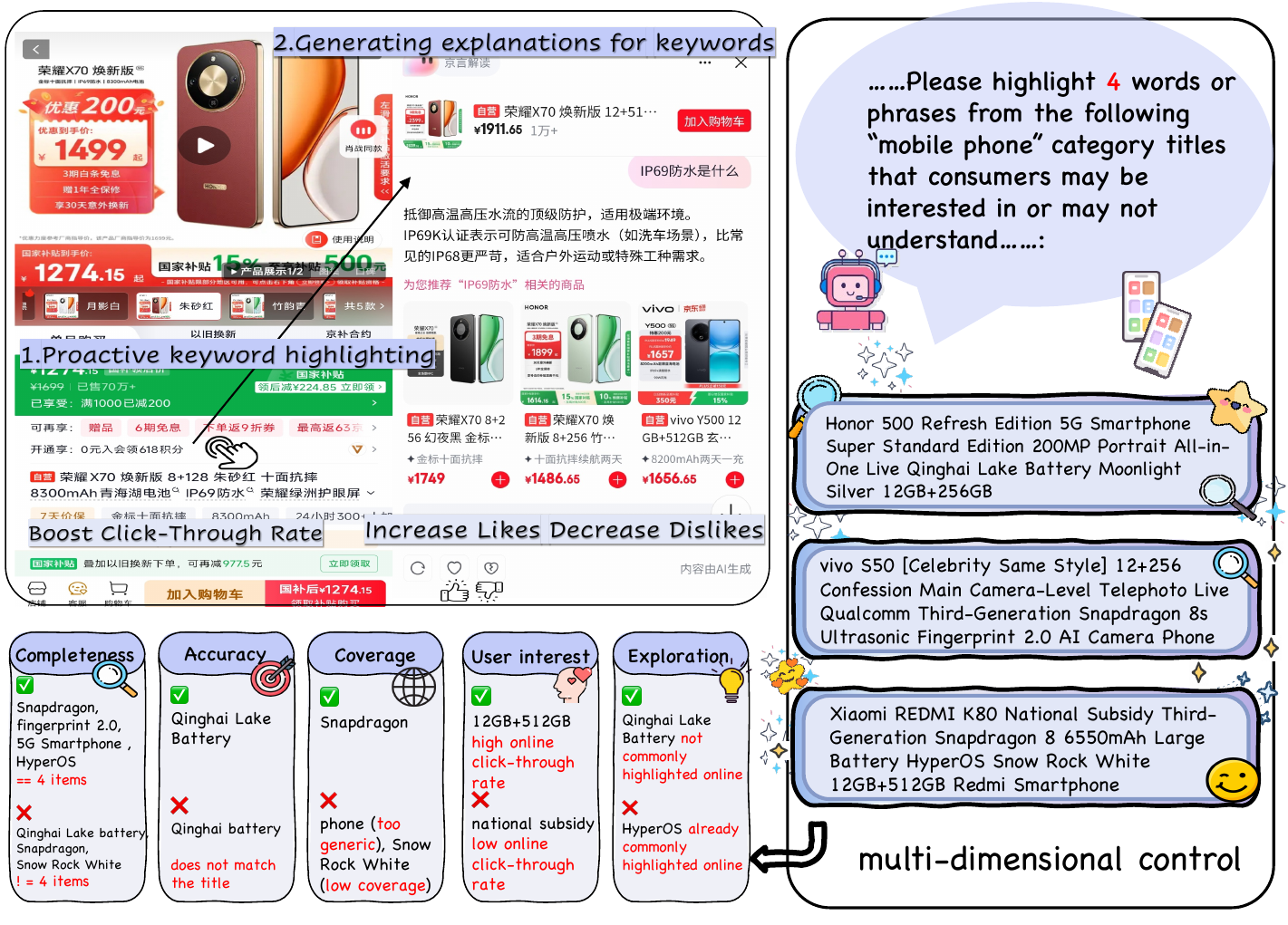} 
\caption{On product detail pages, users can click highlighted keywords to view definitions and provide feedback via likes or dislikes. The online optimization objective is to boost user interest by driving higher click-through rates (CTR) and likes, while minimizing dislikes and maximizing the discovery of novel, previously unmentioned terms. The offline objective aims to enhance extraction quality in terms of completeness, accuracy, and coverage.}
\label{intro}
\end{figure}

\section{Introduction}

E-commerce product detail pages often suffer from high information density. Users typically focus on specific professional terms that require significant cognitive effort to digest. To improve both textual comprehension and user experience, we investigate a joint task of ``term extraction and explanation generation''  (Figure~\ref{intro}), where the model extracts informative terms from product descriptions and generates concise explanations for them. Given the product details, the model is required to extract high-value terms and provide explanations, aiming to concurrently enhance offline quality metrics and online user preference metrics.

Unlike traditional text tasks, the primary challenge lies in stable optimization within a MORL framework. This manifests in three aspects: \textbf{First}, user feedback is inherently sparse and skewed; directly utilizing raw clicks or likes leads to reward degradation. \textbf{Second}, systematic conflicts exist among objectives (e.g., coverage, click-through rate, and like ratios often dictate divergent gradients), where simple scalarization severely blurs training signals. \textbf{Third}, a ``reward tug-of-war'' frequently occurs late in training, causing metrics to oscillate or regress as dominant objectives overshadow weaker ones.

To overcome these dilemmas, we eschew coarse global compromises in favor of MMPO. As a novel alignment framework, MMPO applies bottom-up, customized interventions, effectively crafting a tailored whip to drive each specific objective. The framework consists of three core modules:
(1) At the \textbf{data  and reward level}, we independently reshape and smooth reward signals via exposure stratification and debiasing to enhance their learnability. 
(2) At the \textbf{optimization level}, we orthogonally decouple multi-objective gradients within a low-dimensional subspace to isolate systematic conflicts. 
(3) At the \textbf{dynamic constraint level}, we design a self-prompted restriction strategy to adaptively suppress overly dominant dimensions during late-stage training, effectively mitigating performance oscillations.

The main contributions of this paper are summarized as follows:
\begin{itemize}
\item We formulate the e-commerce term extraction task as a MORL problem with sparse and conflicting rewards, proposing an ``independent-objective-first'' philosophy for real-world industrial deployments.
    
 \item We propose \textbf{MMPO}, integrating reward smoothing, gradient decoupling, and adaptive restriction to efficiently resolve systematic conflicts and late-stage metric oscillations.
    
  \item Extensive experiments show MMPO consistently improves offline completeness and online business metrics in real-world e-commerce. It also delivers competitive performance on general RL tasks such as ToolRL and code generation.

\end{itemize}

\section{Related Work}

Addressing objective conflicts is a core challenge in MORL. Existing methods can be roughly divided into three categories. The first category relies on static reward shaping or global scalarization strategies, using pre-defined weights to balance multiple objectives \citep{cho2025ars, wang2025mhf, kim2026fairdice, shakerinava2026beyond, cai2023distributional}. While simple to implement, these methods typically use fixed weight configurations, making them difficult to adapt to changing conflict patterns across different training stages. As a result, they often lack flexibility when facing stage-dependent conflicts. The second category alleviates objective conflicts through gradient projection or gradient regularization \citep{kim2025conflictaverse, byeon2026multiobjective, yang2025preference, Wu2025ImbalancedGI, 2021Imbalanced}. These methods mainly handle local gradient conflicts at the current iteration, but they usually lack holistic modeling and control of training dynamics, making it difficult to maintain stable coordination among multiple objectives over long training trajectories. The third category introduces reference models, trust regions, or constrained optimization mechanisms to limit the magnitude of policy updates \citep{yuan2025moduli, chen2026a, xu2026metaalignerbidirectionalpreferencepolicyoptimization}. However, these constraints are typically pre-specified and fixed, and therefore cannot be adapted to different optimization stages. Consequently, they often struggle to balance exploration, stability, and multi-objective trade-offs during early training, convergence, or periods of intensified conflicts.

In large language model alignment, GRPO \citep{shao2024deepseekmathpushinglimitsmathematical} performs optimization by taking a weighted sum of advantages, but this can cause different reward terms to compete with each other during aggregation, thereby exacerbating reward conflicts and the "tug-of-war" phenomenon. GDPO \citep{liu2026gdpogrouprewarddecouplednormalization} alleviates reward scale inconsistency and reward tug-of-war by independently normalizing each reward before aggregation. Although this approach partially reduces reward conflicts, it still remains at the level of reward aggregation and does not further consider the coordination among data distribution, gradient propagation, and constraint mechanisms during training. In contrast to simple adjustments at the advantage level, our framework performs full-chain coordinated correction across data, gradient, and constraint dimensions, providing a more systematic solution to signal loss, gradient contamination, and training instability in complex multi-objective scenarios.

\section{Method} 
\begin{figure*}[t] 
\centering
\includegraphics[width=0.9\textwidth]{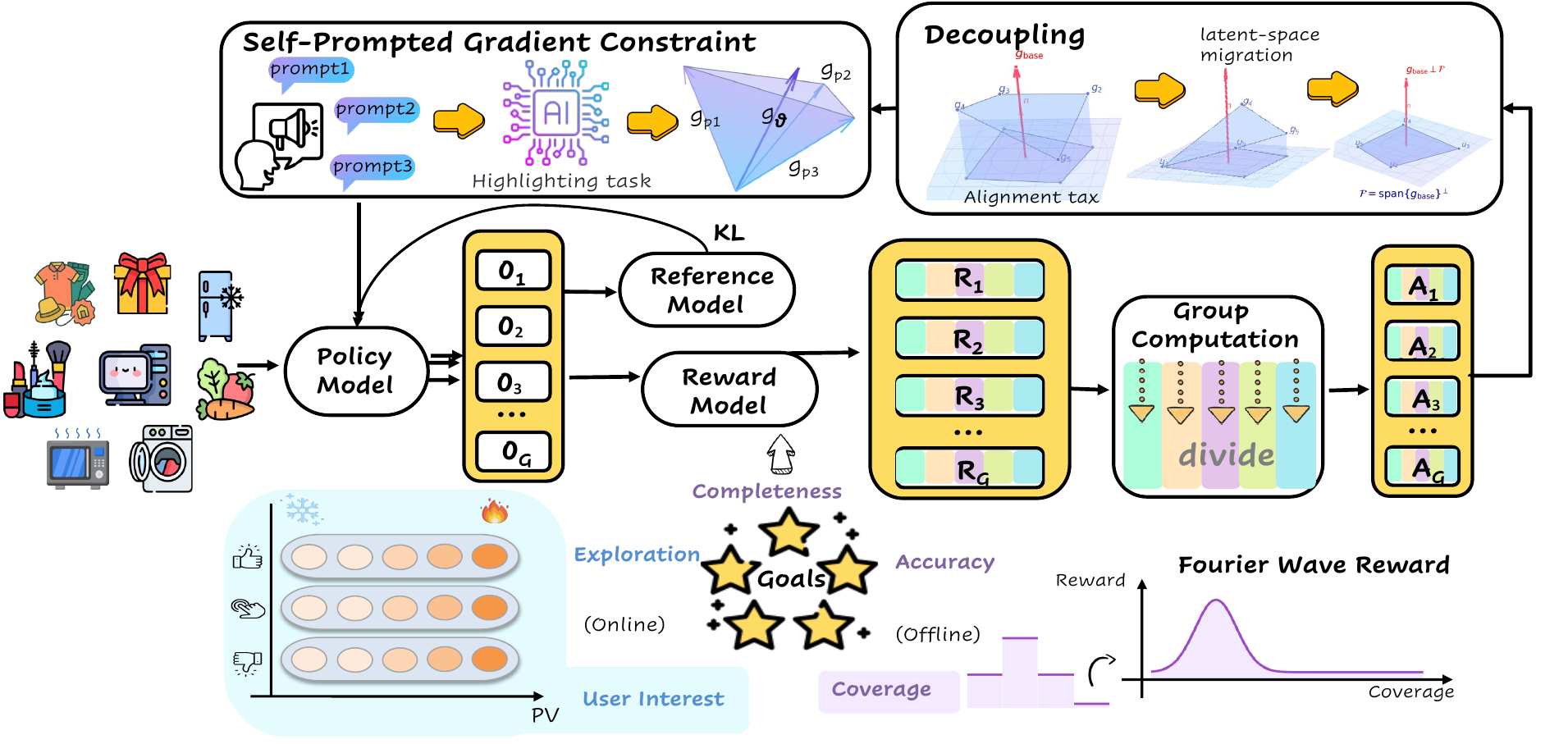}
\caption{Our method begins by inputting a sequence of product titles into the policy model to generate candidate outputs. To balance multiple optimization objectives, we first refine the reward signals by applying exposure-based bias correction to online feedback (likes, dislikes, and clicks). To address reward sparsity, we utilize Fourier-based reward smoothing to enhance the granularity and discriminative power of the reward signals. Regarding objective conflicts, we decompose the rewards into distinct dimensions during the gradient update phase. Furthermore, we employ a hybrid constraint mechanism, integrating standard KL divergence with self-prompted gradient constraints, to regulate policy updates and ensure training stability.}
\label{fig2}
\end{figure*}

Given a set of product descriptions $x$ for a specific category, our goal is to learn a word selection policy that extracts key segments $s_i$ and generates their corresponding explanations $e_i$, resulting in the set $\{(s_i, e_i)\}_{i=1}^{n}$. We formulate this task as a sequential decision-making process driven by multi-faceted reward signals. Beyond maintaining offline quality---specifically completeness, precision, and coverage---the policy must also adapt to online user feedback, such as clicks and likes, to maximize user acceptance. Since these objectives are often misaligned and do not improve monotonically, the task is fundamentally a Multi-Objective Reinforcement Learning problem rather than a simple generation task.

This industrial scenario presents three primary challenges:
\textbf{Reward Sparsity:} The lack of stable feedback for most candidate segments often leads the model to over-exploit high-frequency words, hindering the exploration of novel, high-value ones. Without direct user signals, rewards tend to degrade into weak offline metrics, failing to provide sustained guidance for preference optimization.
\textbf{Reward Conflict:} Objectives such as increasing coverage and suppressing dislikes often pull the model in different directions. Simple scalarization of these rewards can cause one dimension to dominate, thereby compromising others.
\textbf{Reward Tug-of-War:} As the policy converges, it tends to gravitate toward high-weight reward dimensions. This leads to training instability, regression in local metrics, or even reward collapse.
These hurdles represent the core difficulties of optimizing reinforcement learning in real-world business contexts.

\subsection{Data and Rewards}

We define five reward dimensions with detailed descriptions provided in Figure~\ref{intro}.

\paragraph{Bias Correction in User Feedback.}
For the user interest dimension, we utilize click-through rates (CTR), likes, and dislikes from the past three months as reference signals. However, these signals are inherently biased: high-exposure categories (e.g., smartphones) generate abundant feedback, while niche categories (e.g., ceremonial items) often show near-zero signals despite having high-quality segments. To resolve this, we first construct a candidate pool by merging product titles, category lexicons, and rule-based extractions. We then apply exposure-based stratification and bias correction to normalize feedback across different categories and items. This pipeline identifies ``what to learn'' and ``which signals are reliable,'' thereby determining the resolution and the practical upper bound of the user-interest reward.

\paragraph{Fourier Smoothing for Sparse Rewards.}
To mitigate reward sparsity---especially for the coverage signal---we introduce a Fourier-based smoothing mechanism inspired by coordinate-based representation learning ~\citep{tancik2020fourier}. Specifically, we apply Fourier feature mapping to the input layer of the reward model, thereby projecting discrete segment representations into a multi-band sinusoidal feature space. This enables the reward model to learn a smoother mapping from segment features to reward estimates. By adjusting the frequency bandwidth, isolated reward observations can be generalized into a smoother reward landscape, allowing similar or adjacent segments to receive progressive feedback rather than purely binary signals, while still preserving the model's ability to capture local high-frequency peaks, i.e., segments with exceptionally high user interest.

\subsection{Decoupling}

\subsubsection{Reward Conflict}
In the practice of multi-objective alignment for Large Language Models (LLMs), a common approach is naive linear gradient summation:
\begin{equation}
g_{\text{sum}} = \sum_{i=1}^{K} w_i g_i,
\end{equation}
where $g_i$ denotes the gradient of the $i$-th objective and $w_i$ is its corresponding weight. However, such a direct aggregation often introduces a significant alignment tax, mainly due to directional interference among gradients.

Suppose we have a gradient $g_{\text{base}}$ representing fundamental constraints (e.g., accuracy and completeness) and a gradient $g_{\text{pref}}$ representing user preferences (e.g., exploration and user interest). In the non-linear parameter space, a direct summation may introduce components of $g_{\text{pref}}$ that conflict with $g_{\text{base}}$, thereby interfering with the model's core capabilities during updates. To mitigate this issue, we aim to constrain the update direction to a safe subspace that minimizes interference with the base gradients, rather than enforcing an overly strict orthogonality condition.

\subsubsection{Objective Decomposition}
To make the above idea computationally tractable, we transform implicit gradient interference into explicit projection-based updates. We decompose any preference gradient $g_{\text{pref}}$ into two orthogonal components:
\begin{equation}
g_{\text{pref}} = g_{\text{pref}}^{\parallel} + g_{\text{pref}}^{\perp},
\end{equation}
where $g_{\text{pref}}^{\parallel}$ denotes the component aligned with the base-gradient subspace and $g_{\text{pref}}^{\perp}$ denotes the remaining component in the feasible tangent space $\mathcal{F}$.

\textbf{Proposition 1 (First-Order Projection Consistency).}
If the update direction $d$ is restricted to the safe region $\mathcal{F}$, then the directional contribution of the original preference gradient is equivalent to that of its projected component:
\begin{equation}
g_{\text{pref}}^\top d
=
\left(g_{\text{pref}}^{\parallel} + g_{\text{pref}}^{\perp}\right)^\top d
=
\left(g_{\text{pref}}^{\perp}\right)^\top d,
\end{equation}
where the equality holds because the component $g_{\text{pref}}^{\parallel}$ lies outside the feasible update space and does not contribute to the update within $\mathcal{F}$. In other words, within the safe update region, conflicting gradient components can be removed without losing valid preference-alignment signals. The core objective of MMPO is to reconstruct raw gradient summation into such purified projections.

\subsubsection{Priority-Aware Sequential Projection}
Although the projection-based formulation is conceptually appealing, constructing projection operators in the full parameter space $\mathbb{R}^d$ is computationally expensive. In practice, we observe that the effective update information is concentrated in a low-dimensional subspace:
\begin{equation}
S = \text{span}\{g_1, \dots, g_K\}.
\end{equation}
Therefore, instead of operating in the full parameter space, we perform a change of basis within this subspace.

We introduce a \textbf{Priority-Aware Sequential Projection} mechanism. The key idea is to combine reward decoupling with an explicit priority order: base capabilities are assigned the highest priority, and their gradients are used to establish the primary basis directions of the subspace. The remaining preference objectives are then sequentially orthogonalized with respect to the already established directions. This process consists of two steps:

\begin{itemize}
    \item \textbf{Interference Stripping}: Each gradient $g_k$ is transformed into a basis vector $u_k$ by removing its projections onto all preceding higher-priority directions. The resulting basis vectors are normalized so that $U^\top U = I$, where $U = [u_1, \dots, u_K]$.
    \item \textbf{Reparameterized Control}: Once the orthonormal basis $U$ is obtained, the aggregated gradient is reparameterized into independent coordinates:
    \begin{equation}
    c = U^\top g_{\text{sum}}, \qquad g_{\text{update}} = Uc.
    \end{equation}
\end{itemize}

Gradient decomposition acts as a prerequisite for intervention: by mapping entangled gradients into a transparent coordinate system, it provides the ‘surgical field’ required for priority-aware projection. This synergy between transformation and intervention bridges reward decoupling with multi-objective optimization.

At its core, this design re-parameterizes the space, allowing preference objectives to navigate around base-task trajectories. Conflicts are thus treated as redundant information—inherently filtered out by constraining preference gradients to a compatible subspace. This eliminates interference at the source, ensuring precise alignment without complex, ad-hoc interventions.

\subsection{Self-Prompted Gradient Constraint}

While the decomposition mechanism in Section 3.2 effectively promotes multi-dimensional rewards during early training, late-stage optimization often encounters a ``reward tug-of-war'' dominated by high-weight objectives. In this regime, a specific reward dimension may continue to surge while others decline, leading to policy oscillations or even local collapse. To mitigate this, we propose the Self-Prompted Gradient Constraint. Rather than relying on heuristic weight adjustments, Self-Prompted Gradient Constraint leverages the model's inherent instruction-following capabilities to define an ``endogenous boundary,'' establishing a safe interval for gradient updates.

Specifically, for each task, we construct augmented prompts $\{x_{\text{aug}}^{(d)}\}_{d=1}^{3}$ across three critical dimensions: \textit{completeness}, \textit{accuracy}, and \textit{coverage}. These prompts incorporate explicit reinforcement instructions (e.g., ``Ensure the output strictly adheres to the $N$-word length limit and maintains structural integrity''), eliciting the model's strongest instruction-aligned gradient responses $g_{\text{aug}}^{(d)}$ under ideal conditions. We then project these gradients into the subspace $S$ to derive a set of reference coordinates:
\begin{equation}
\mathcal{C}_{\text{ref}} = \left\{c_{\text{aug}}^{(d)} \mid c_{\text{aug}}^{(d)} = U^\top g_{\text{aug}}^{(d)}, \ d=1,2,3 \right\}.
\end{equation}
The minimum and maximum values of these reference coordinates along each axis define a coordinate envelope.

To achieve adaptive update constraints, we introduce a coordinate clamping mechanism. For the original coordinate components $c$ computed from the current update, we enforce them to remain within the interval defined by the self-prompted signals:
\begin{equation}
\hat{c}_k = \text{clip}\left(c_k, \min_d(c_{\text{aug},k}^{(d)}), \max_d(c_{\text{aug},k}^{(d)})\right).
\end{equation}
The final policy update gradient is reconstructed as
\begin{equation}
g_{\mathrm{final}} = U \hat{c}.
\end{equation}
This approach functions as an internal ``navigation limiter'': the three dimensions of self-prompted signals delineate a multi-dimensional convex hull on the gradient manifold, forming an endogenous trust region. When a specific preference dimension attempts to push the model beyond its cognitive baseline due to overfitting, Self-Prompted Gradient Constraint forcibly pulls the update back to the compliant boundary via coordinate clamping. By utilizing this multi-dimensional redundancy check, the model retains its potential for complex preference alignment while anchoring its core capabilities within a robust cognitive scope, effectively mitigating reward oscillations and local collapse issues.

\section{Experiments}

\paragraph{Experimental Setup.}
All training and inference were conducted on NVIDIA B200 GPUs. For the reinforcement learning phase, the rollout size (group size per prompt) was set to 8.

\paragraph{Benchmark Selection.}
For our internal product-detail span selection scenario, the training set consists of 65,232 products and the evaluation set contains 8,879 products. To model user interest feedback, we further collected user behavior data from the past three months across various product categories. These data were utilized as supervision signals during training to enhance the model’s sensitivity to real-world business preferences.
Regarding external evaluation, we selected a representative set of recent public benchmarks as test datasets. The specific benchmarks used in each sub-experiment are detailed in their respective sections below.

\paragraph{Model Selection.}
We primarily adopted DeepSeek-series~\citep{deepseekai2025deepseekr1incentivizingreasoningcapability} and Qwen-series~\citep{qwen3} models as the main experimental backbones. It is worth noting that in code generation tasks, we found that conventional instruction-tuned models often fail to receive effective training feedback, especially for constraints such as code formatting, syntactic structure, and executability. Therefore, for such experiments, we prioritize dedicated code models to ensure more stable training signals and more learnable rewards.

\paragraph{Baseline Methods.}
We selected GRPO~\citep{shao2024deepseekmathpushinglimitsmathematical}and GDPO~\citep{liu2026gdpogrouprewarddecouplednormalization} as the main baselines. Both methods are representative recent preference optimization approaches and provide a meaningful comparison for multi-objective optimization settings, making them suitable baselines for evaluating the effectiveness of our method.

\subsection{Main Results}

\subsubsection{Offline Evaluation} 


\begin{figure}[t] 
\centering
\includegraphics[width=1.1\columnwidth]{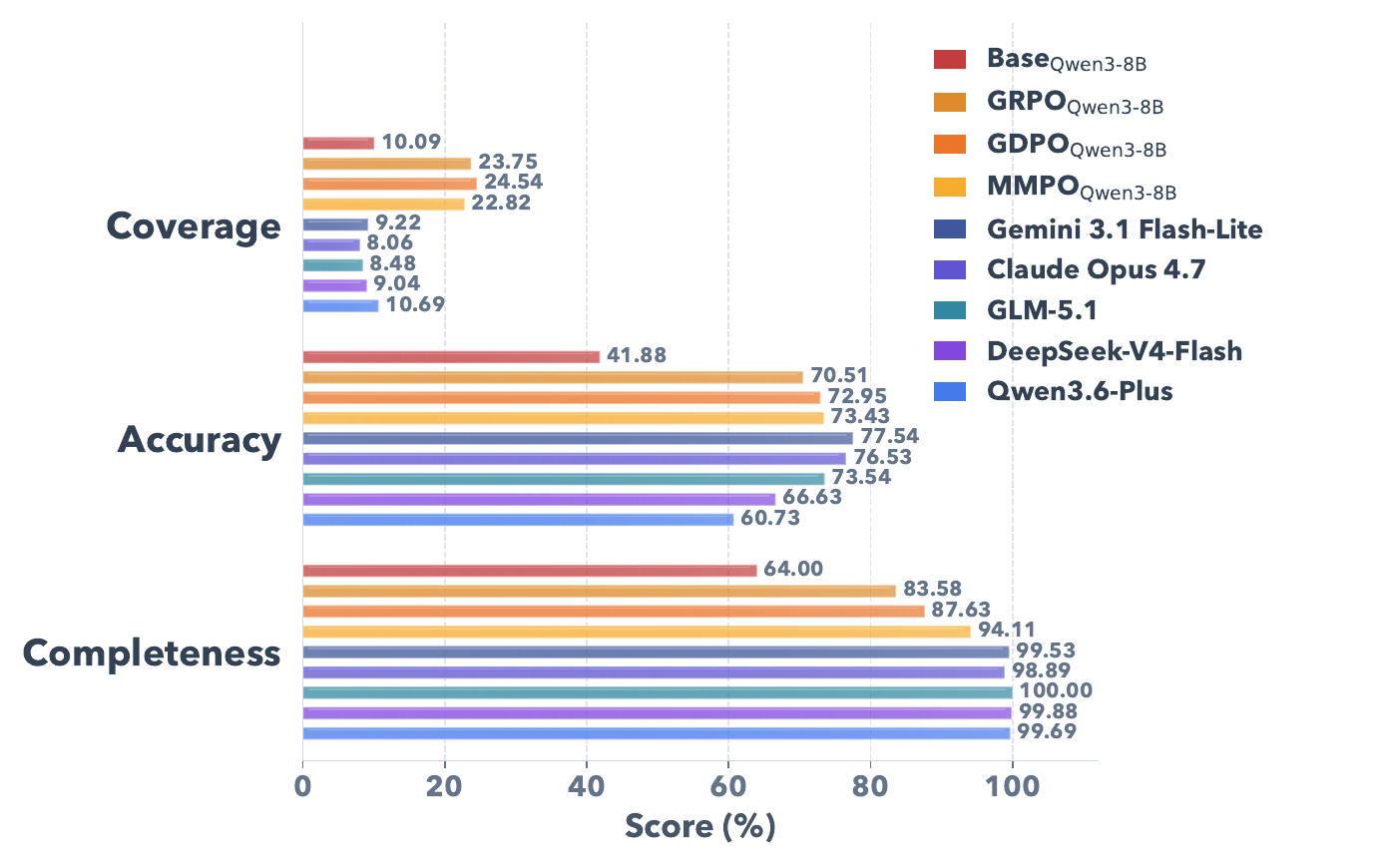} 
\caption{Comparison of different models on three core metrics: Completeness ($\uparrow$), Accuracy ($\uparrow$), and Coverage (optimal at 0.22).}
\label{off_lab}
\end{figure}

As shown in Figure \ref{off_lab}, $\text{MMPO}_{\text{Qwen3-8B}}$ establishes a decisive advantage over all 8B-scale baselines, ranking first on \emph{every} metric. It attains a Completeness of 94.11\% and an Accuracy of 73.43\%, exceeding the strongest baseline $\text{GDPO}_{\text{Qwen3-8B}}$ by 6.48\% on Completeness while further improving Accuracy, and improving over the RL baseline $\text{GRPO}_{\text{Qwen3-8B}}$ by 10.53\% and 2.92\%, respectively. Crucially, on the Coverage metric, MMPO scores 22.82\%---the closest of all methods to the ideal value of $\sim$22\%---whereas both GRPO (23.75\%) and GDPO (24.54\%) overshoot this target. This confirms that our integrated methodology effectively tames the reward dominance and over-generation that plague scalarization- and normalization-based methods, yielding balanced, well-calibrated multi-dimensional convergence rather than trading one objective off against another.

More strikingly, under a \emph{unified evaluation benchmark} (a sampled subset of 1,500 products across 50 categories), $\text{MMPO}_{\text{Qwen3-8B}}$ remains highly competitive with---and on key axes superior to---frontier Large Language Models (LLMs), including 
\textbf{Gemini 3.1 Flash-Lite}~\citep{Gemini_3.1_Flash}, 
\textbf{Claude Opus 4.7}~\citep{TheC3}, 
\textbf{GLM-5.1}~\citep{glm5_1}, 
\textbf{DeepSeek-V4-Flash}~\citep{deepseekai2025deepseekr1incentivizingreasoningcapability}, and 
\textbf{Qwen3.6-Plus}~\citep{qwen36plus}. In terms of Accuracy, $\text{MMPO}_{\text{Qwen3-8B}}$ surpasses DeepSeek-V4-Flash (66.63\%) and Qwen3.6-Plus (60.73\%) and matches GLM-5.1 (73.54\%). Its advantage is most pronounced on Coverage: MMPO's 22.82\% more than doubles the best frontier model (10.69\%), as every general-purpose LLM severely under-covers the multi-constraint requirements. These results suggest that the \textbf{MMPO framework}'s synergistic design effectively mitigates the \textit{alignment tax} of multi-objective learning, allowing a specialized lightweight model to narrow—and in some dimensions, bridge—the performance gap with frontier large-scale models.

\begin{figure}[t] 
\centering
\includegraphics[width=0.5\textwidth]{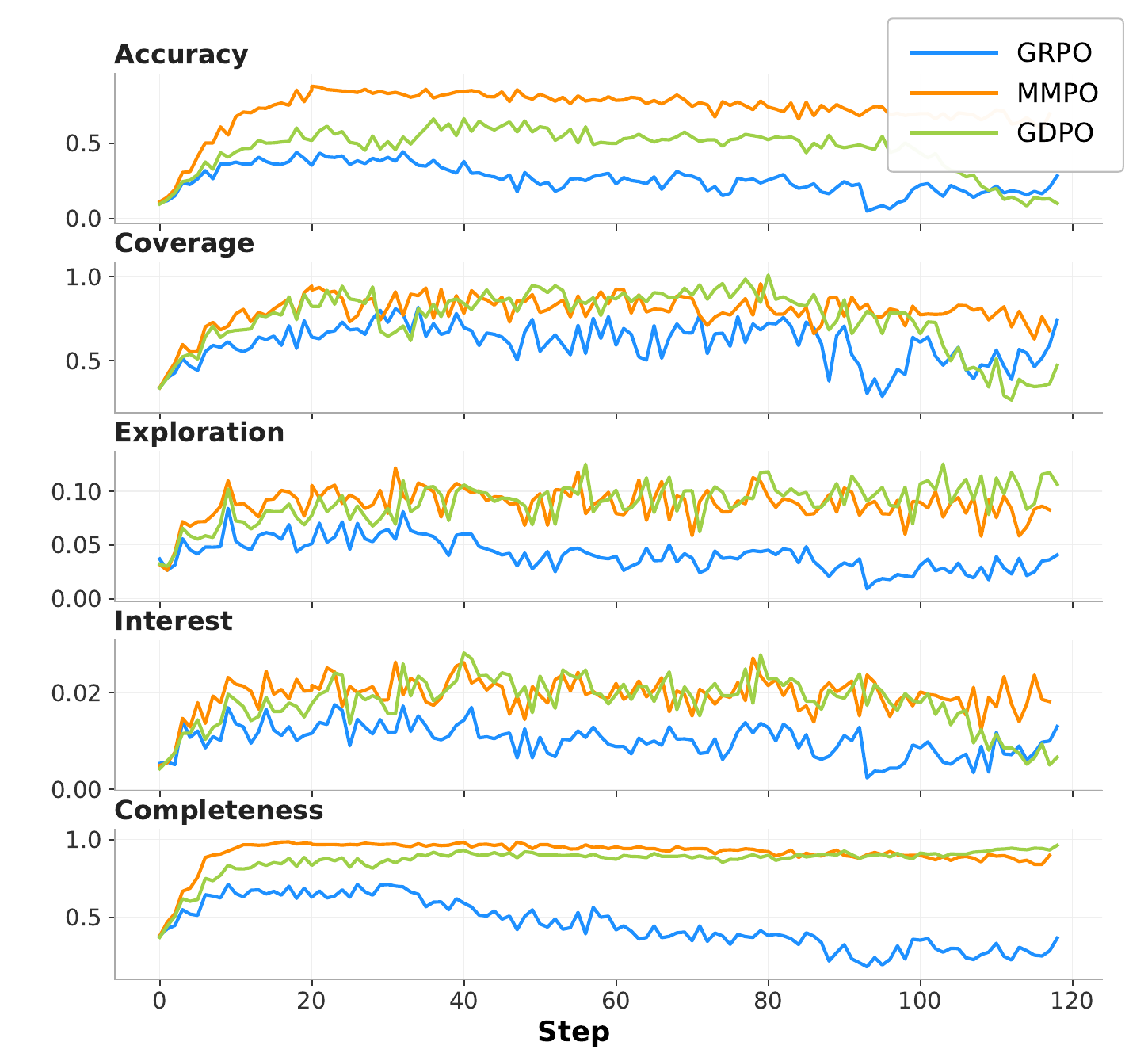} 
\caption{Training dynamics of the five core reward metrics.}
\vspace{-10pt}
\label{fig_reward}
\end{figure}

As shown in Figure \ref{fig_reward}, we use a $1 \times 10^{-6}$ learning rate and a $256$ batch size for stable convergence, regulated by a $0.001$ KL penalty to prevent distribution drift. For efficiency, vLLM-based rollouts are employed with a $0.6$ GPU memory limit, balancing generation throughput and training requirements. Unlike GRPO~\citep{shao2024deepseekmathpushinglimitsmathematical} and GDPO~\citep{liu2026gdpogrouprewarddecouplednormalization}, which exhibit frequent fluctuations and severe metric collapse (e.g., $\text{Accuracy}$ and $\text{Completeness}$) in later training stages, MMPO (orange curve) ensures faster convergence and exceptional stability. This robustness is attributed to the subspace orthogonal decoupling mechanism, which prevents ``reward hacking'' caused by objective conflicts. Ultimately, MMPO simultaneously optimizes core metrics and multi-dimensional constraints like $\text{Exploration}$ and $\text{Interest}$, ensuring comprehensive synergistic alignment throughout training and evaluation.

\subsubsection{A/B Testing}

 To validate MMPO in production, we conducted a two-week online A/B test with 10\% traffic, comparing an MMPO-fine-tuned Qwen3-8B model against a DeepSeek-V3 baseline.
As shown in Table~\ref{tab:ab_main_results}, the MMPO-based model consistently outperformed the control, achieving a 5.07\% increase in Entry-click GMV, a 3.14\% increase in Entry-click Orders, and gains in Unique Visitors (+2.96\%), UV Value (+1.77\%), Total Interaction Turns (+2.54\%), and Active Users (+1.99\%). All improvements are statistically significant at the 95\% confidence level (p<0.01p<0.01), confirming that our offline performance gains translate effectively into real-world business value.

\begin{table}[htbp]
\centering
\small
\setlength{\tabcolsep}{0.5pt}
\begin{tabular}{lccc}
\toprule
Metric & Relative Change & 95\% CI & $p$-value \\ 
\midrule
Entry-click GMV & +5.07\% & [3.21\%, 6.93\%] & $<0.001$ \\
Entry-click Orders & +3.14\% & [1.82\%, 4.46\%] & $<0.001$ \\
Unique Visitors & +2.96\% & [1.53\%, 4.39\%] & $<0.001$ \\
Total Interaction Turns & +2.54\% & [1.18\%, 3.90\%] & $<0.001$ \\
Active Users & +1.99\% & [0.71\%, 3.27\%] & $0.002$ \\
Overall UV Value & +1.77\% & [0.52\%, 3.02\%] & $0.006$ \\
\bottomrule
\end{tabular}
\caption{Relative improvements of the treatment group over the control group in the online A/B test.}
\label{tab:ab_main_results}
\end{table}

\subsection{Ablation Results}

\begin{table}[htbp]
\centering

\setlength{\tabcolsep}{3pt}
\begin{tabular}{lccc}
\toprule
\textbf{Method} & \textbf{Comp.} & \textbf{Acc.} & \textbf{Cov.} \\
\midrule
GRPO (w/ SR) & 59.63\% & 62.48\% & 22.51\% \\
\quad w/o SR  & 3.52\%  & 4.10\%  & 6.88\%  \\
\quad w/ SD   & 63.62\% & 61.22\% & 22.71\% \\
\quad w/ SP   & 90.16\% & 72.81\% & 22.52\% \\
\textbf{MMPO (SD+SP)} & \textbf{93.37\%} & \textbf{74.62\%} & \textbf{22.12\%} \\
\bottomrule
\end{tabular}
\caption{Ablation study results on a subset of 1,000 products. The evaluated components include Smoothed Reward (SR), Subspace Decoupling (SD), and Self-Prompt Gradient Limitation (SP). Completeness and Accuracy are higher-is-better, while Coverage is closer to 0.22 when better. Note that SR is applied to the GRPO baseline by default to mitigate reward sparsity.To ensure the robustness of our results, we performed a post-hoc analysis by partitioning the test set into five disjoint subsets; the measured variance across these subsets is negligible (standard deviation < 0.35\% for all metrics), confirming the stability of our findings.}
\label{tab:performance}
\end{table}

\begin{table*}[htbp]
\centering

\resizebox{\textwidth}{!}{%
\begin{tabular}{ll ccccccccc}
\toprule
\multirow{2}{*}{\textbf{Model}} & \multirow{2}{*}{\textbf{Method}} 
& \textbf{Live} & \textbf{Multi-turn} & \textbf{Non-live} 
& \cellcolor{myblue!30}\textbf{Avg} & \cellcolor{myblue!30}\textbf{Correct} 
& \textbf{Lv1} & \textbf{Lv2} & \textbf{Lv3} & \cellcolor{myblue!30}\textbf{Overall} \\
& & \textbf{Acc} & \textbf{Acc} & \textbf{Acc} 
& \cellcolor{myblue!30}\textbf{Acc} & \cellcolor{myblue!30}\textbf{Format}
& \textbf{Acc} & \textbf{Acc} & \textbf{Acc} & \cellcolor{myblue!30}\textbf{Acc} \\
\midrule

\multirow{4}{*}{\textbf{Qwen3-8B}} 
& Base & 53.10\% & 65.31\% & 71.34\% & \cellcolor{myblue!30}{63.25\%} & \cellcolor{myblue!30}{94.81\%} & 69.92\% & 55.22\% & 43.51\% & \cellcolor{myblue!30}{62.48\%} \\
& GRPO & 55.86\% & 72.45\% & \textbf{77.52\%} & \cellcolor{myblue!30}{68.61\%} & \cellcolor{myblue!30}{97.32\%} & 74.94\% & 64.18\% & 44.27\% & \cellcolor{myblue!30}{67.00\%}  \\
& GDPO & 56.90\%  & \textbf{73.21\%} & 76.55\% & \cellcolor{myblue!30}{68.89\%} & \cellcolor{myblue!30}{98.21\%} & \textbf{75.44\%} & \textbf{65.67\%} & 41.98\% & \cellcolor{myblue!30}{67.00\%}  \\
& MMPO & \textbf{57.93\%} & 72.96\% & 77.21\% & \cellcolor{myblue!30}{\textbf{69.36\%}} & \cellcolor{myblue!30}{\textbf{98.83\%}} & \textbf{75.44\%} & 64.18\% & \textbf{46.56\%} & \cellcolor{myblue!30}{\textbf{67.84\%}} \\
\midrule

\multirow{4}{*}{\textbf{Qwen3-4B}} 
& Base & 53.10\% &62.24\% & 67.75\% & \cellcolor{myblue!30}{61.03\%} & \cellcolor{myblue!30}{92.29\%} & 65.91\% & 53.73\% &48.09\% & \cellcolor{myblue!30}{60.64\%} \\
& GRPO & \textbf{56.55}\% & 67.60\%  & 72.96\% & \cellcolor{myblue!30}{65.70\%}  & \cellcolor{myblue!30}{97.54\%} & 70.43\% & 56.72\% & \textbf{49.62\%} & \cellcolor{myblue!30}{64.32\%} \\
& GDPO & 54.14\% & 67.35\% & 72.31\% & \cellcolor{myblue!30}{64.60\%}  & \cellcolor{myblue!30}{99.33\%} & 70.68\% & 53.73\% & 46.56\% & \cellcolor{myblue!30}{63.48\%} \\
& MMPO & 54.83\% & \textbf{68.37\%} & \textbf{75.24\%} & \cellcolor{myblue!30}{\textbf{66.15\%}} & \cellcolor{myblue!30}{\textbf{99.83}\%} & \textbf{71.18\%} & \textbf{61.19}\% & \textbf{49.62\%} & \cellcolor{myblue!30}{\textbf{65.33}\%} \\
\midrule

\multirow{4}{*}{\textbf{DS-R1-Distill-Qwen-7B}} 
& Base & \textbf{37.24\%} & 34.44\% & 29.64\% & \cellcolor{myblue!30}{33.77\%} & \cellcolor{myblue!30}{32.66\%} & 38.10\% & 17.91\% & 26.72\% & \cellcolor{myblue!30}{33.33\%} \\
& GRPO & 35.52\% & 29.85\% & 26.38\% & \cellcolor{myblue!30}{30.58\%} & \cellcolor{myblue!30}{34.34\%} & 33.08\% & 19.40\% & 29.77\% & \cellcolor{myblue!30}{30.82\%} \\
& GDPO & \graycell{37.24\%} & \graycell{34.44\%} & \graycell{29.64\%} & \cellcolor{myblue!30}{\graycell{33.77\%}} & \cellcolor{myblue!30}{\graycell{32.66\%}} & \graycell{38.1\%} & \graycell{17.91\%} & \graycell{26.72\%} & \cellcolor{myblue!30}{\graycell{33.33\%}} \\
& MMPO & 36.90\%  & \textbf{41.33\%} & \textbf{49.84\%} & \cellcolor{myblue!30}{\textbf{42.69\%}} & \cellcolor{myblue!30}{\textbf{36.34\%}} & \textbf{50.38\%} & \textbf{22.39\%} & \textbf{33.59\%} & \cellcolor{myblue!30}{\textbf{43.55\%}} \\
\midrule

\multirow{4}{*}{\textbf{DS-R1-Distill-Qwen-1.5B}} 
& Base & 12.76\% & 9.18\% & 6.19\% & \cellcolor{myblue!30}{9.38\%} & \cellcolor{myblue!30}{29.48\%} & 9.52\% & 10.45\% & 8.40\% & \cellcolor{myblue!30}{9.38\%} \\
& GRPO & \textcolor{gray}{12.76\%} & \textcolor{gray}{9.18\%} & \textcolor{gray}{6.19\%} & \cellcolor{myblue!30}{\textcolor{gray}{9.38\%}} & \cellcolor{myblue!30}{\textcolor{gray}{29.48\%}} & \textcolor{gray}{9.52\%} & \textcolor{gray}{10.45\%} & \textcolor{gray}{8.40\%} & \cellcolor{myblue!30}{\textcolor{gray}{9.38\%}} \\
& GDPO & \textcolor{gray}{12.76\%} & \textcolor{gray}{9.18\%} & \textcolor{gray}{6.19\%} & \cellcolor{myblue!30}{\textcolor{gray}{9.38\%}} & \cellcolor{myblue!30}{\textcolor{gray}{29.48\%}} & \textcolor{gray}{9.52\%} & \textcolor{gray}{10.45\%} & \textcolor{gray}{8.40\%} & \cellcolor{myblue!30}{\textcolor{gray}{9.38\%}} \\
& MMPO & \textbf{14.48\%} & \textbf{20.92\%} & \textbf{26.71\%} & \cellcolor{myblue!30}{\textbf{20.7\%}} & \cellcolor{myblue!30}{\textbf{36.53\%}} & \textbf{26.32\%} & \textbf{14.93\%} & \textbf{8.87\%} & \cellcolor{myblue!30}{\textbf{19.53\%}} \\
\bottomrule
\end{tabular}
}
\caption{Performance Comparison of Different Models and Methods on ToolRL}
\label{tab:model_comparison}
\end{table*}

Table \ref{tab:performance} validates the core components of MMPO. Removing Smoothed Reward (w/o SR) causes both completeness and accuracy to plummet below 5\%, demonstrating that SR is essential for learning under extreme reward sparsity. Furthermore, both Subspace Decoupling (SD) and Self-Prompt Gradient Limitation (SP) independently contribute to performance gains. Notably, SP drives a substantial increase in completeness (59.63\% $\rightarrow$ 90.16\%) and accuracy (62.48\% $\rightarrow$ 72.81\%), effectively preventing reward dominance in later training stages. Ultimately, the full MMPO framework (SD + SP) achieves the best results (93.37\% completeness, 74.62\% accuracy), highlighting the strong synergy between low-level gradient decoupling and high-level prompt-based constraints in resolving multi-objective conflicts.

\subsection{General Capabilities}

\subsubsection{ToolRL Evaluation}

We evaluate our approach using the ToolRL \citep{qian2026toolrl} benchmark, assessing performance across three dimensions: difficulty levels (Lv1–3 and overall accuracy), interaction scenarios (real-time, multi-turn, and offline), and output format adherence. Our training framework employs a multi-objective reward function comprising: \textbf{Normalized Accuracy} (to be maximized), \textbf{Format compliance} (to be maximized), and \textbf{Output Length} (to be optimized for appropriate reasoning depth), which collectively empower the model's final performance. 


As shown in Table \ref{tab:model_comparison}, MMPO achieves state-of-the-art performance among Qwen-based models, particularly in terms of average accuracy (Avg Acc). A notable observation is the behavior of DeepSeek-R1-Distill models, where severe reward sparsity—stemming from a mismatch between pre-training preferences and the evaluation reward function—causes GRPO~\citep{shao2024deepseekmathpushinglimitsmathematical} and GDPO~\citep{liu2026gdpogrouprewarddecouplednormalization} to stall or degrade due to a lack of feedback. In contrast, MMPO exhibits superior robustness, boosting the Overall Acc of DS-7B from 33.33\% to 43.55\%. While MMPO shows a marginal performance gap in Live Acc compared to baseline methods on some models , it maintains the highest overall Avg Acc. This underscores MMPO’s effectiveness in global optimization, avoiding overfitting to specific local metrics while ensuring comprehensive multi-objective alignment.

\subsubsection{Code Generation}
We evaluate MMPO on LiveCodeBench~\citep{jain2024livecodebench} (DeepSeek-Coder-1.3B~\citep{deepseek-coder} / Qwen2.5-Coder-3B~\citep{hui2024qwen2}) using $\text{pass@}k$ ($k \in \{1, 5, 10, 20\}, n=20$) to assess accuracy, diversity, and search robustness. Our training framework employs a multi-objective reward function, comprising: \textbf{Pass} (higher is better) to maximize correctness; \textbf{Exceed} (lower is better) to minimize time-limit violations; \textbf{Bug} (lower is better) to reduce syntax and logical errors; and \textbf{Length} (optimized for appropriate reasoning depth) to regulate output scale. 

\begin{figure}[t] 
\centering
\includegraphics[width=1.0\columnwidth]{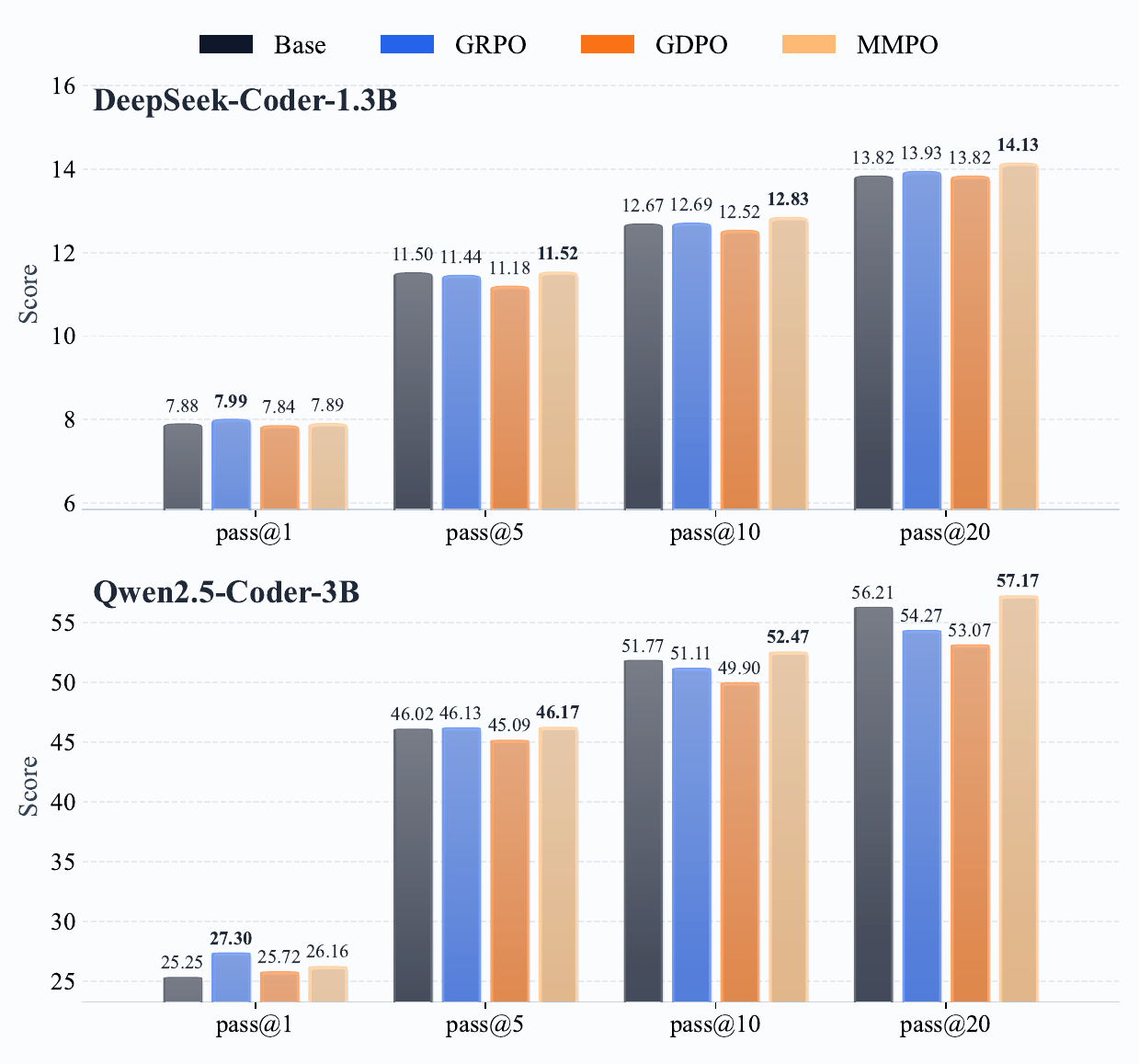} 
\caption{Code Generation Performance (pass@k) Comparison.}
\label{pass_k1}
 \vspace{-4pt}
\end{figure}

As shown in Figure~\ref{pass_k1}, while greedy baselines (GRPO~\citep{shao2024deepseekmathpushinglimitsmathematical}, GDPO~\citep{liu2026gdpogrouprewarddecouplednormalization}) achieve competitive $\text{pass@}1$ scores by exploiting dominant rewards, they severely suffer from mode collapse. Conversely, MMPO harmonizes the accuracy-diversity trade-off. It maintains strong $\text{pass@}1$ accuracy while dominating higher-$k$ metrics, effectively avoiding local optima to broaden solution coverage. Notably, MMPO reaches 57.17\% $\text{pass@}20$ on Qwen2.5-Coder-3B, demonstrating superior multi-objective alignment and generalizability.

\vspace{-5pt}


 \section{Conclusion}

We introduced MMPO to address sparse rewards, gradient conflicts, and metric oscillations in Multi-Objective Reinforcement Learning. By leveraging exposure debiasing, subspace orthogonal decoupling, and self-prompted constraints, MMPO achieves stable, multi-faceted LLM alignment without reward collapse. Empirical results across real-world e-commerce applications and standard benchmarks (ToolRL, code generation) demonstrate that MMPO significantly outperforms baselines like GRPO~\citep{shao2024deepseekmathpushinglimitsmathematical} and GDPO~\citep{liu2026gdpogrouprewarddecouplednormalization}, establishing a robust and scalable framework for complex alignment tasks involving multi-objective tradeoffs.

\section*{Limitations}

While MMPO achieves significant gains, it inherently inherits limitations from underlying LLMs. First, standardizing the evaluation of “process fidelity” remains an open challenge ~\citep{iclr_shu2026dare}. Second, foundation models still exhibit critical vulnerabilities to advanced threats, including multimodal “adversarial smuggling” ~\citep{iclr_li2026makingmllmsblindadversarial} and multi-turn strategic red-teaming ~\citep{iclr_guo2026treebased}. These bottlenecks are universal to current foundation models and evaluation ecosystems rather than flaws in our framework. Future work will integrate robust defense mechanisms and process monitoring to improve reliability in adversarial environments.

\newpage
\clearpage

\bibliography{custom}

\end{document}